\pdfoutput=1

\documentclass[11pt]{article}

\usepackage{acl}

\usepackage{times}
\usepackage{latexsym}
\usepackage{booktabs}
\usepackage{multirow}
\usepackage{colortbl}
\usepackage{pifont}
\usepackage[T1]{fontenc}
\usepackage{xurl}

\usepackage[utf8]{inputenc}

\usepackage{microtype}
\usepackage{graphicx}
\usepackage{amsmath, amsfonts}
\usepackage{bbm}
\usepackage{subcaption}

\title{Prompting ELECTRA: Few-Shot Learning with \\ Discriminative Pre-Trained Models}

\newcommand{\affilsup}[1]{\rlap{\textsuperscript{\normalfont#1}}}
\author{
    Mengzhou Xia\affilsup{1}
    \qquad Mikel Artetxe\affilsup{2}
    \qquad Jingfei Du\affilsup{2} \\
    \textbf{Danqi Chen}\affilsup{1}
    \qquad \textbf{Ves Stoyanov}\affilsup{2} \\
    $^1$Princeton University \quad $^2$Meta AI\\
    \texttt{\{mengzhou, danqic\}@cs.princeton.edu} \\
    \texttt{\{artetxe, jingfeidu, ves\}@meta.com} \\
}

\newcommand{\mask}{[\texttt{MASK}]}

\newcommand{\cmark}{\ding{51}}

\newcommand\tf[1]{\textbf{#1}}

\newcommand{\CLS}{\texttt{[CLS]}}

\definecolor{maroon}{cmyk}{0,0.87,0.68,0.32}
\definecolor{ggreen}{HTML}{d4e7cf}
\definecolor{rred}{HTML}{f1c6c6}
\definecolor{yyellow}{HTML}{fff014}
\begin{document}
\maketitle
\begin{abstract}

Pre-trained masked language models successfully perform few-shot learning by formulating downstream tasks as text infilling. However, as a strong alternative in full-shot settings, discriminative pre-trained models like ELECTRA do not fit into the paradigm. In this work, we adapt prompt-based few-shot learning to ELECTRA and show that it outperforms masked language models in a wide range of tasks. ELECTRA is pre-trained to distinguish if a token is generated or original. We naturally extend that to prompt-based few-shot learning by training to score the originality of the target options without introducing new parameters. Our method can be easily adapted to tasks involving multi-token predictions without extra computation overhead. Analysis shows that ELECTRA learns distributions that align better with downstream tasks.\footnote{Code is available at \url{https://github.com/facebookresearch/ELECTRA-Fewshot-Learning}.}

\end{abstract}


\begin{figure}[t!]
\centering
\includegraphics[width=\linewidth]{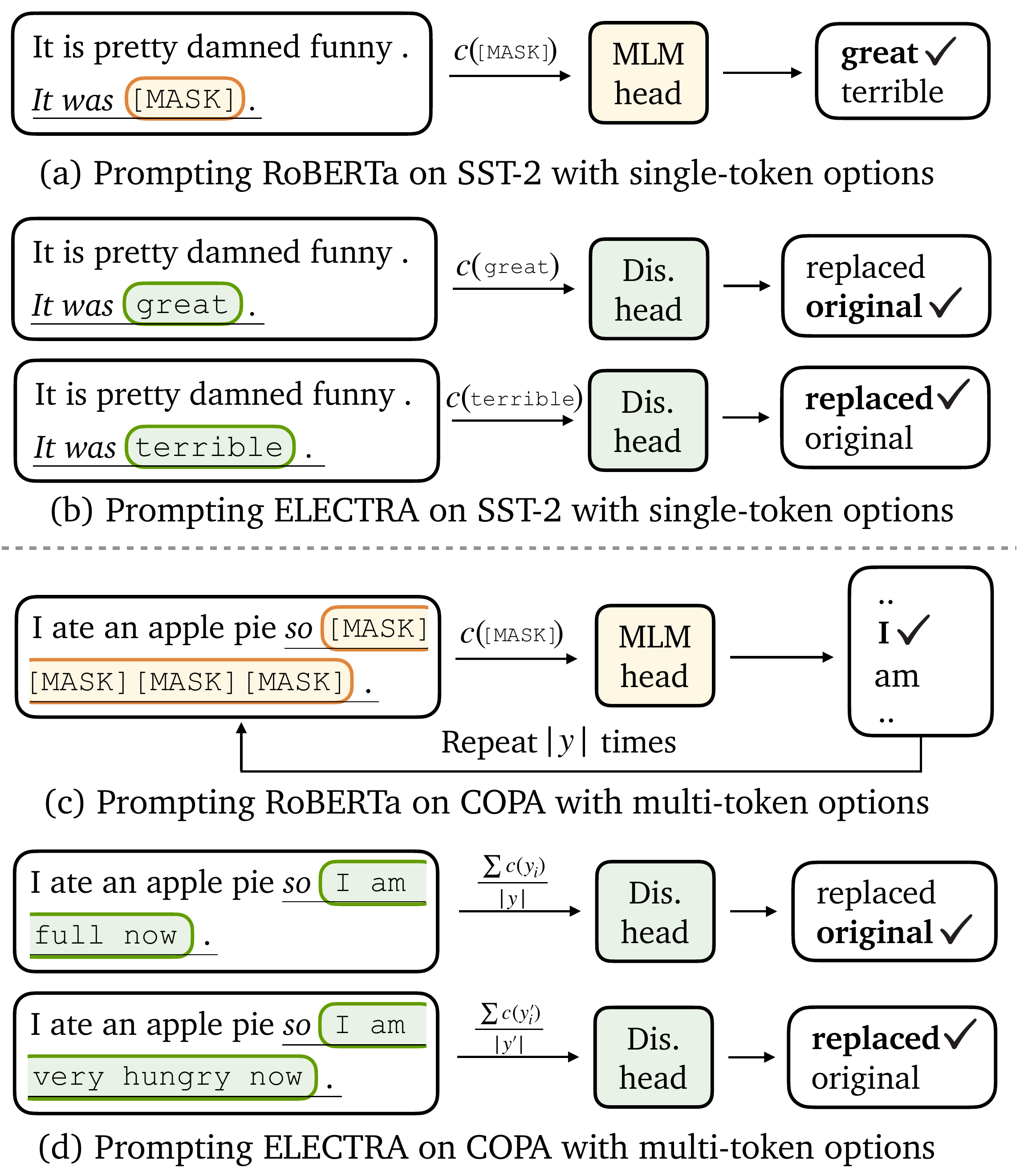}
\caption{Prompt-based fine-tuning with RoBERTa and ELECTRA for two downstream tasks: SST-2~\cite{socher2013recursive} and COPA~\cite{roemmele2011choice}. The underlined text is the task-specific {template}. $c(\cdot)$: contextualized embedding; $y$ and $y'$: a correct and an incorrect option, respectively. }

\label{fig:teaser}
\end{figure}
\section{Introduction}

Large pre-trained language models are known to be effective zero-shot and few-shot learners when scaled~\cite{brown2020language, artetxe2021efficient, rae2021scaling}. Much smaller masked language models (MLMs), like BERT~\cite{devlin2019bert} and RoBERTa~\cite{liu2019roberta}, can be fined-tuned with only a few examples by utilizing prompt-based fine-tuning, which updates the model to select the correct target word or option~\cite{schick2021exploiting, gao2021making}.

In this paper, we hypothesize that discriminative pre-trained models like ELECTRA \cite{Clark2020ELECTRA:} will make even stronger few-shot learners as alternatives to MLMs as they are pre-trained to distinguish between challenging alternatives.
To test this hypothesis, we explore prompt-based learning with ELECTRA by aligning its pre-training objective---distinguishing if a single token is generated or original---with prompt-based predictions for downstream tasks. We reuse ELECTRA's discriminative head to classify the correct target word as original tokens. As an additional benefit, we can naturally adapt the approach to multi-token spans by aggregating either hidden representations or output probabilities. In contrast, MLMs require autoregressive decoding to adapt to multi-token options \cite{schick2021s}.

We propose an approach to prompting ELECTRA, as shown in \autoref{fig:teaser}. Though trained with the same or even less computation than BERT and RoBERTa, ELECTRA turns out to be a more effective few-shot learner. It outperforms BERT and RoBERTa by 10.2 and 3.1 points on average across 9 tasks with single-token options for base-sized models in the few-shot setting, and the trend prevails for large-sized models. ELECTRA also outperforms RoBERTa on 4 tasks with multi-token options. Our analysis suggests that the failing predictions from ELECTRA’s generator could actually feed negatives with opposite meanings from the correct tokens to the discriminator, which strengthens ELECTRA’s ability to distinguish concepts with opposite meanings for zero-shot predictions.

\section{Background}
\subsection{Prompting Masked Language Models}
MLMs such as BERT and RoBERTa are trained by masking words in inputs and maximizing the probability of original tokens that are replaced by \mask~tokens. Given a sequence $x_1, x_2, \cdots, x_n$ with the $i$-th token masked, the objective is:
\begin{equation*}
- \log \frac{\exp{\left(c({\mask}) \cdot \mathbf{e}_{x_i}\right)}}{\sum_{v \in \mathcal{V}} \exp{\left(c({\mask}) \cdot \mathbf{e}_{v}\right)}},
\end{equation*}
where ${\mathbf{e}}_{v}$ denotes the embedding of the word ${v} \in \mathcal{V}$. We use $c(\cdot)$ to denote the contextualized representation for simplicity. Prompt-based learning turns the objective into a softmax distribution over all the target words of a prompt template \cite{ schick2021exploiting,gao2021making}. For example, in binary sentiment analysis, given an input sentence $x$, its associated label $y \in \{\text{positive}, \text{negative}\}$ and a template $\mathcal{T}$, we formulate the prompt as:
\begin{gather}
    \mathcal{T}(x) = x~\text{It was} \texttt{\mask}~. \nonumber
\end{gather}

\noindent By defining a mapping $\mathcal{M}: \mathcal{Y} \rightarrow \mathcal{V}$ from the task label space to words in the vocabulary, the task is transformed into predicting the target word ${\mathcal{M}(y)}$:
\begin{equation*}
    -\log \frac{\exp \left(\mathrm{c}{(\mask)} \cdot \mathbf{e}_{\mathcal{M}(y)}\right)}{\sum_{y' \in \mathcal{Y}}\exp \left(\mathrm{c}{(\mask)} \cdot \mathbf{e}_{\mathcal{M}(y')}\right)}.
\end{equation*}

This formulation can be used for prompt-based zero-shot evaluation and few-shot fine-tuning to perform gradient updates. For tasks involving multi-token options, such as multiple-choice tasks, prompt-based fine-tuning with MLMs is less intuitive. For example, \citet{schick2021s} adopt a multi-class hinge loss for training and devise a heuristic decoding method to estimate probabilities for target options during inference. The disadvantages are (1) such usage of MLMs deviates from the pre-training objective; (2) the pseudo-autoregressive decoding approach cannot forward in batches during inference, which is computationally inefficient.

\subsection{Discriminative Pre-trained Models}

Discriminative pre-trained models such as ELECTRA~\cite{Clark2020ELECTRA:} cast the word prediction problem into a binary classification problem. In ELECTRA, a discriminator and a smaller generator are jointly trained with the goal to distinguish if the tokens are sampled from the generator or from the original data:

\vspace{-1em}
\begin{align}
  - \sum_{i} & \left( \mathbbm{1}(x_i' = x_i)\log \mathcal{H}(c(x_i)) \right. \nonumber \\
  & \left. + \mathbbm{1}(x_i' \neq x_i) \log (1-\mathcal{H}(c{(x_i')}) \right) \nonumber, 
\end{align}
where $\{x_i\}$ are tokens from the original sentence, $\{x_i'\}$ are tokens from the corrupted sentence, and $\mathcal{H}$ denotes the discriminator head. We refer readers to \citet{Clark2020ELECTRA:} for more details.
\begin{table*}[t!]
\centering
\resizebox{2.0\columnwidth}{!}{%
\begin{tabular}{lccc|ccc|ccc} \toprule
                     & \multicolumn{3}{c}{\tf{SST-2}}            & \multicolumn{3}{c}{\tf{SST-5}}             & \multicolumn{3}{c}{\tf{MR}}               \\  \midrule
                     & BERT       & RoBERTa    & ELECTRA    & BERT       & RoBERTa     & ELECTRA    & BERT       & RoBERTa    & ELECTRA    \\ \cmidrule(l{2pt}r{2pt}){2-4} \cmidrule(l{2pt}r{2pt}){5-7}\cmidrule(l{2pt}r{2pt}){8-10}
 \cellcolor{ggreen!30}Zero-shot (\cmark) & 61.6     & 77.8       & \textbf{82.8}       & 26.0       & 30.3        & \textbf{31.1}       & 55.8       & 77.7       & \textbf{81.5}       \\
\cellcolor{ggreen!60}Few-shot & 72.8 (6.4) & \textbf{84.5 (2.3)} & 78.2 (7.6) & 34.9 (2.0) & 37.9 (1.3)  & \textbf{41.7 (1.8)} & 70.8 (5.2) & \textbf{76.8 (3.7)} & 76.3 (2.9) \\
\cellcolor{ggreen!60}Few-shot (\cmark)  & 84.6 (1.0) & 89.9 (0.6) & \textbf{91.2 (0.7)} & 37.9 (1.4) & 43.3 (1.2)  & \textbf{49.3 (1.5)} & 78.2 (1.1) & 85.0 (0.9) & \textbf{88.0 (0.5)} \\
\midrule
\cellcolor{ggreen!90}Full-shot    & 93.2 (0.3) & 94.8 (0.3) & \textbf{{ 95.5 (0.1) }} & 53.4 (0.1) & \textbf{{ 55.8 (0.1) }} & 54.8 (0.2) &  86.8 (0.3) & 88.7 (0.2) & \textbf{{ 90.3 (0.0) }}       \\ \midrule
                     & \multicolumn{3}{c}{\tf{MNLI}}             & \multicolumn{3}{c}{\tf{RTE}}               & \multicolumn{3}{c}{\tf{QNLI}}             \\ \midrule
                     & BERT       & RoBERTa    & ELECTRA    & BERT       & RoBERTa     & ELECTRA    & BERT       & RoBERTa    & ELECTRA    \\ \cmidrule(l{2pt}r{2pt}){2-4} \cmidrule(l{2pt}r{2pt}){5-7}\cmidrule(l{2pt}r{2pt}){8-10}
\cellcolor{ggreen!30}Zero-shot (\cmark)    & 43.5       & 48.1       & \textbf{51.9}       & 48.7       & 53.4        & \textbf{57.8}       & 49.5       & 50.5       & \textbf{54.5}       \\
\cellcolor{ggreen!60}Few-shot & 41.3 (1.7) & 42.2 (2.8) & \textbf{44.7 (3.1)} & 52.8 (4.1) & 54.2 (2.8)  & \textbf{59.1 (1.7)} & 68.4 (4.8) & 65.1 (5.1) & \textbf{69.7 (3.7)} \\
\cellcolor{ggreen!60}Few-shot (\cmark)  & 47.9 (0.7) & 59.1 (2.1) & \textbf{60.8 (2.3)} & 57.5 (2.6) & 62.7 (2.2) & \textbf{67.0 (1.4)} & 56.0 (0.7) & 67.4 (2.8) & \textbf{70.6 (4.0)} \\
\midrule
\cellcolor{ggreen!90}Full-shot     & 84.7 (0.3) & { 87.4 (0.0) } & \textbf{{ 88.6 (0.0) }} & { 69.1 (1.6) } & { 74.2 (0.2) } & \textbf{{ 78.3 (1.1) }} & { 91.6 (0.1) } & { 92.6 (0.1) } & \textbf{{ 93.2 (0.0) }}       \\ \midrule
                     & \multicolumn{3}{c}{\tf{SNLI}}             & \multicolumn{3}{c}{\tf{AGNews}}            & \multicolumn{3}{c}{\tf{BoolQ}}            \\ \midrule
                     & BERT       & RoBERTa    & ELECTRA    & BERT       & RoBERTa     & ELECTRA    & BERT       & RoBERTa    & ELECTRA    \\ \cmidrule(l{2pt}r{2pt}){2-4} \cmidrule(l{2pt}r{2pt}){5-7}\cmidrule(l{2pt}r{2pt}){8-10}
\cellcolor{ggreen!30}Zero-shot (\cmark)     & 38.7       & 48.8       & \textbf{56.6}       & 60.6       & \textbf{73.2}        & 72.2       & 47.7       & 55.9       & \textbf{59.1}       \\
\cellcolor{ggreen!60}Few-shot & 50.4 (2.8) & 44.8 (3.9) & \textbf{50.5 (3.3)} & 84.9 (0.6) & \textbf{85.5 (0.8)}  & 81.4 (1.4) & 54.7 (2.5) & 56.8 (3.9) & \textbf{57.2 (2.1)} \\
\cellcolor{ggreen!60}Few-shot (\cmark)  & 51.0 (2.6) & 66.3 (3.0) & \textbf{72.4 (2.0)} & 84.6 (1.2) & \textbf{87.1 (0.6)}  & 86.9 (1.0) & 57.4 (2.9) & 57.8 (2.4) & \textbf{60.8 (4.2)} \\
\midrule
\cellcolor{ggreen!90}Full-shot     & 91.2 (0.0) & 91.9 (0.1) & \textbf{{ 92.4 (0.1) }} & 94.9 (0.0) & \textbf{{ 95.4 (0.1) }} & 94.9 (0.0) & { 75.3 (1.9) } & { 78.6 (0.3) } & \textbf{{ 81.1 (1.1) }}  \\ \bottomrule
\end{tabular}
}
\caption{\colorbox{ggreen!20}{Zero-shot}, \colorbox{ggreen!50}{few-shot} (16 examples per label) and \colorbox{ggreen!80}{full-shot} results of BERT, RoBERTa and ELECTRA base models. \cmark~denotes whether a prompt is used or not (Appendix~\ref{app:template}); otherwise, it adopts standard fine-tuning using the \CLS token. We report average accuracy across 3 runs with standard deviations in parenthesis. We highlight the best number for each setting in bold.}
\label{tab:main}
\end{table*}

\section{Method: Prompting ELECTRA}
Discriminative models like ELECTRA are strong alternatives to MLMs, so they have the potential to be effective few-shot learners even though they do not fit the current paradigm. Furthermore, ELECTRA could be more amenable to solving tasks involving multi-token options by reusing the discriminative head. In this section, we propose adapting ELECTRA to accommodate a wide range of tasks involving single-token or multi-token options for prompt-based learning.\footnote{Two concurrent works explore similar ideas to prompt discriminative pre-trained models \cite{yao-etal-2022-prompt, li2022pre}. Both approaches concatenate the labels in one forward pass while we forward the input with different target words/options. We also demonstrate the effectiveness on multiple-choice tasks.
}

\subsection{Tasks with Single-token Target Words}

The prompts for ELECTRA models are formulated with an input sentence $x$,  a label $y \in \mathcal{Y}$, and a template $\mathcal{T}$ with the mapping function $\mathcal{M}$. An example of sentiment classification is as follows:
\begin{align}
\vspace{-1em}
\mathcal{T}(x, y) = x ~\text{It was}~\mathcal{M}(y)~. \nonumber
\end{align}
For each input sentence, we create $|\mathcal{Y}|$ prompts and forward them for gradient updates such that the model predicts the correct target word as an original token and incorrect ones as generated tokens:

\vspace{-1.5em}
\begin{align}
& - \log \mathcal{H}(c(\mathcal{M}(y))) \nonumber \\ & -\sum\limits_{y' \in \mathcal{Y}/\{y\}}\log (1- \mathcal{H}(c(\mathcal{M}(y')))). \nonumber
\end{align}

\noindent During inference, the model predicts how likely it is for each target option to fit into the sentence and outputs the most likely one. This approach allows us to perform prompt-based zero-shot prediction and few-shot fine-tuning analogously to the MLM paradigm\footnote{We also experimented with a variation to adapt the discriminative objective for contrastive learning, but the results were not as competitive. Please see \autoref{app:con} for details. }. Note that this approach requires forwarding the input with different target words $|\mathcal{Y}|$ times, which is less efficient than MLMs.

\subsection{Tasks with Multi-token Target Options}
We handily adapt ELECTRA's discriminative objective to accommodate tasks with multi-token options for prompt-based fine-tuning. The mapping $\mathcal{M}: \mathcal{Y}\rightarrow \mathcal{V}^*$ is an identity function  for tasks where the target spans are the options themselves. Consider the multiple-choice task COPA \cite{roemmele2011choice}; given a premise $x$, a template $\mathcal{T}$, and an option $y \in \mathcal{Y}$, we formulate the prompt as:
\begin{equation}
\mathcal{T}(x, y) = x~\text{so/because}~\mathcal{M}(y)~. \nonumber
\end{equation}

\noindent As an option $\mathcal{M}(y)$ contains multiple tokens, we either average the hidden representations of all tokens in $\mathcal{M}(y)$ (equivalent to $y$):
\begin{equation}
\mathcal{H}\left(\frac{1}{|y|} \sum_j c(y_{j})\right) \nonumber;
\end{equation}
\noindent where $y_j$ denotes the $j$-th token of an option $y$, or use the average probability of all tokens in $y$ as the final prediction:
\vspace{-0.5em}
\begin{equation}
\frac{1}{|y|} \sum_j\mathcal{H} (c(y_j)); \nonumber
\end{equation}
or simply take {\CLS} token's probability: $\mathcal{H}(c(\CLS)).$ These approaches fully reuse pre-trained weights of ELECTRA, including the discriminator head, and refrain from autoregressive-style decoding. Similar to PET, we only use them for few-shot fine-tuning due to the discrepancy from pre-training.

\section{Experimental Results}

\subsection{Setup}

We run experiments with released checkpoints of BERT \cite{devlin2019bert}, RoBERTa \cite{liu2019roberta} and ELECTRA \cite{Clark2020ELECTRA:} from the \textit{transformers}~\cite{wolf2019huggingface} library. We use base-sized models unless otherwise specified\footnote{More details on pre-trained models are in \autoref{app:model}.}. For tasks with single-token target words, we conduct prompt-based zero-shot evaluations, as well as standard\footnote{We use the {\CLS} token for prediction in standard fine-tuning, known as head fine-tuning in \citet{le2021many}.} and prompt-based few-shot training for each checkpoint. We evaluate on 9 tasks including SST-2, SST-5, MR, MNLI, RTE, QNLI, SNLI, AGNews, and BoolQ\footnote{BoolQ is licensed under \href{https://creativecommons.org/licenses/by-sa/3.0/legalcode}{CC-BY-SA 3.0}.}. For tasks with multi-token options, we evaluate the few-shot setting. These tasks include COPA, StoryCloze, HellaSwag, and PIQA. Details of datasets (including references) and prompts are in \autoref{app:datasets} and \autoref{app:template}.

For our default experiments, we use $16$ examples per label for single-token tasks and 32 examples for multiple-choice tasks.  We follow \citet{gao2021making} to create a development set with the same size as the training set for model selection and conduct three runs of experiments to mitigate instability issues~\cite{dodge2020fine} for all experiments.\footnote{More training details are in \autoref{app:setup}.}

\begin{figure*}[!t]
     \centering
     \begin{subfigure}[b]{\textwidth}
         \centering
         \includegraphics[width=0.32\textwidth]{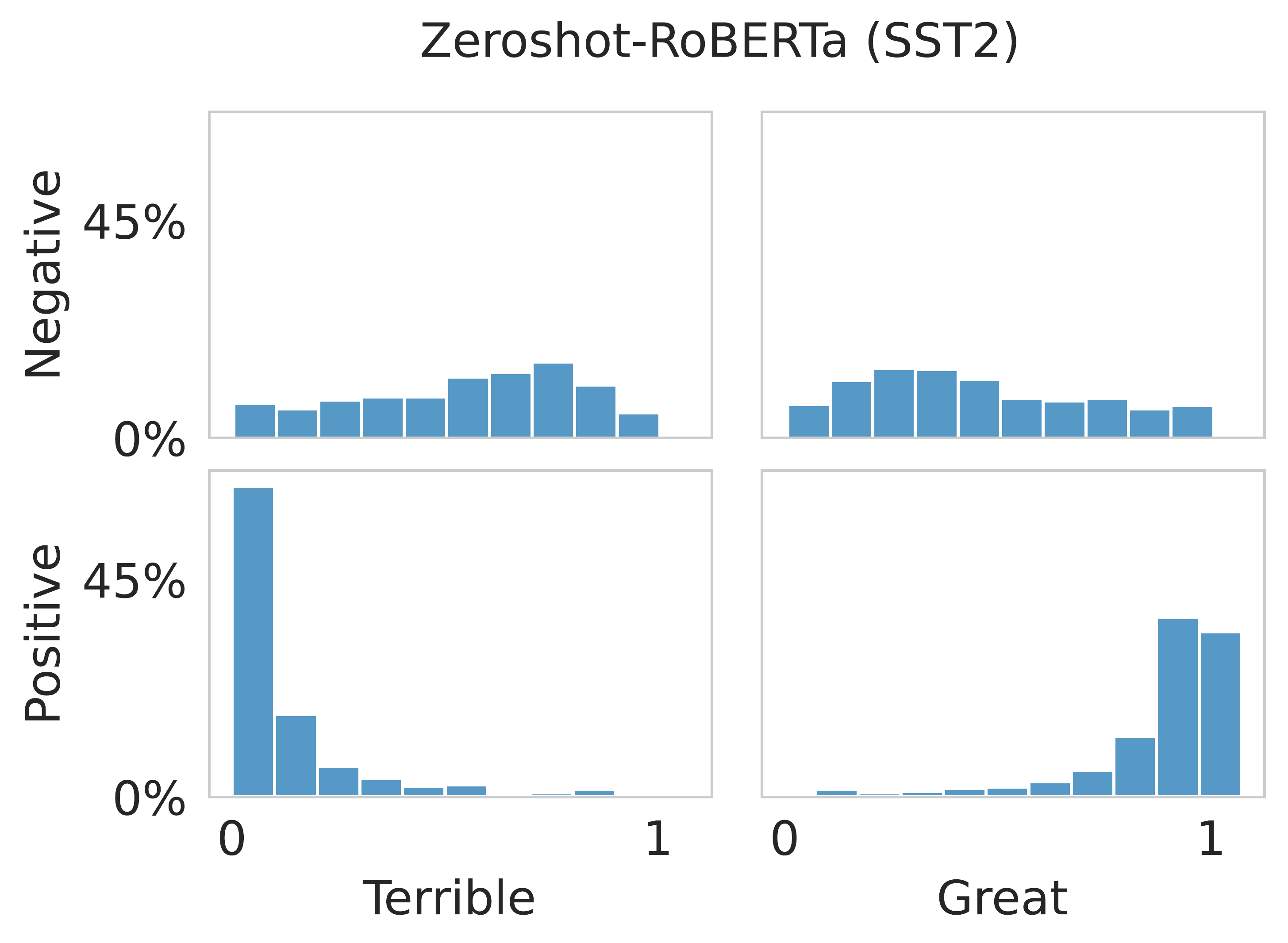}
         \includegraphics[width=0.32\textwidth]{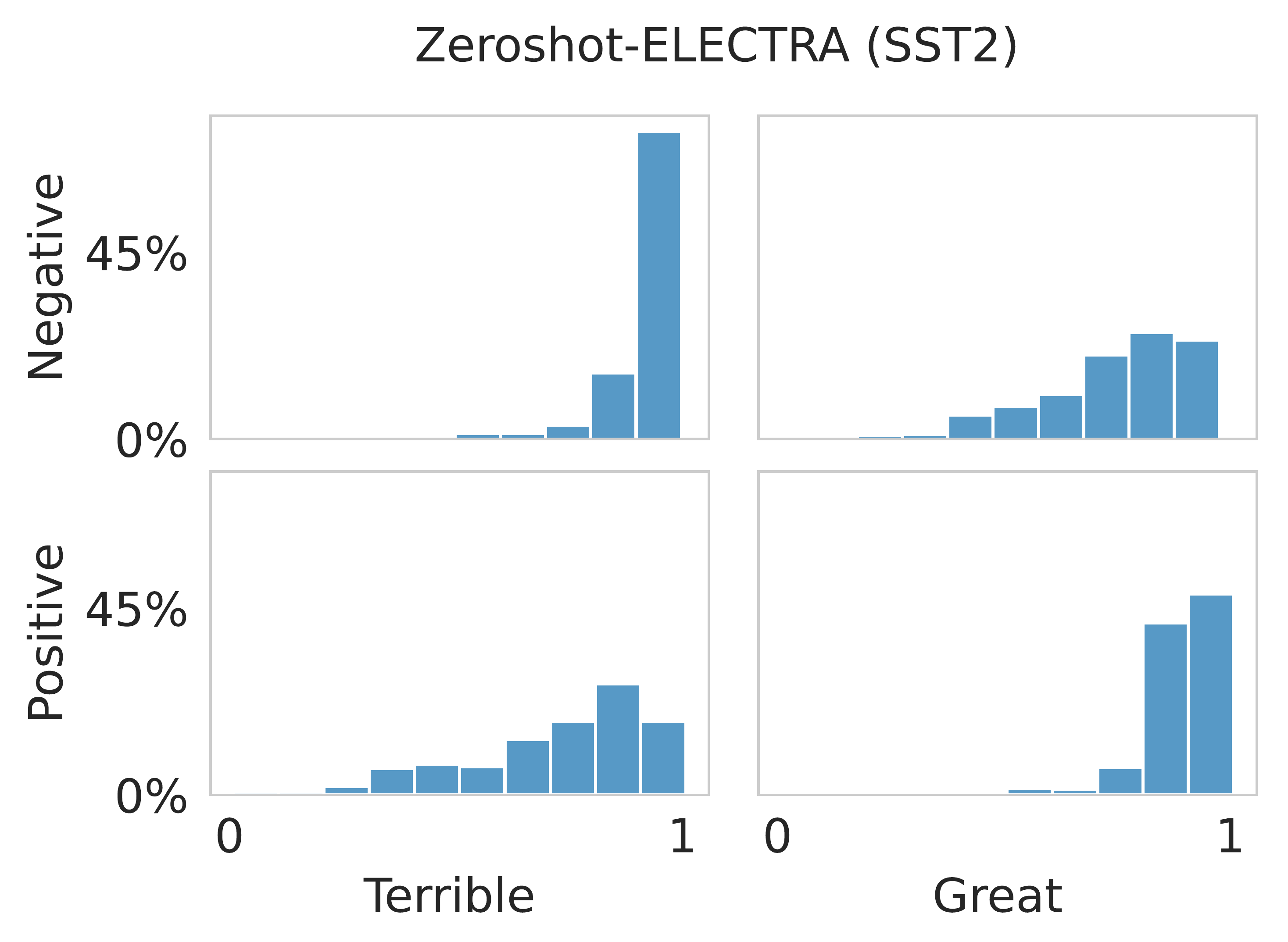}
         \includegraphics[width=0.32\textwidth]{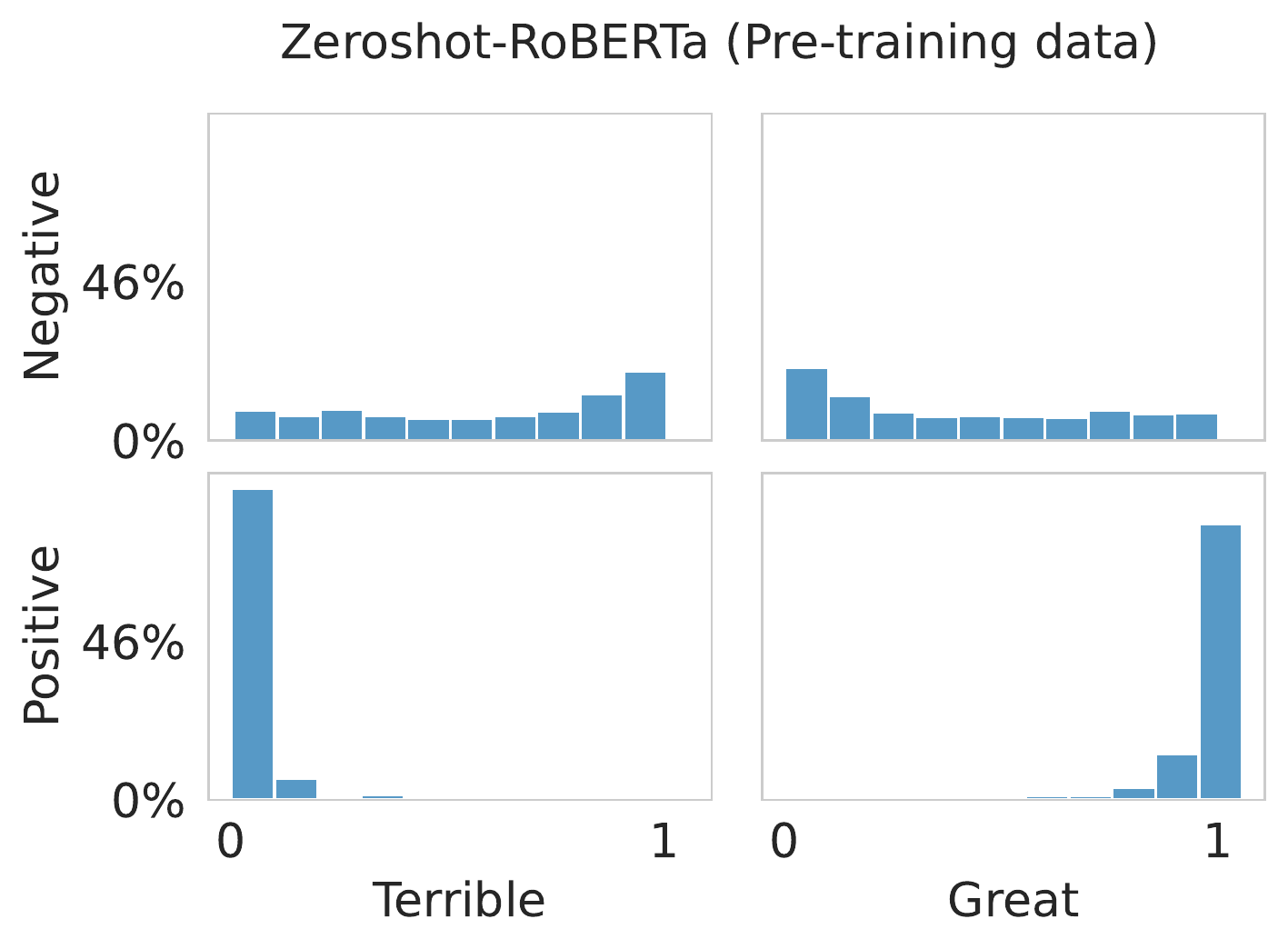}
     \end{subfigure}
     \hfill
     \caption{Zero-shot prediction distributions on SST-2 with RoBERTa (left) and ELECTRA (middle). Zero-shot prediction distributions on pre-training data that contain target words (right). Each sub-graph shows the output distribution for inputs associated with a label $y \in \{\text{negative}, \text{positive}\}$ when prompted with the target words $\{\text{great}, \text{terrible}\}$. The y-axis shows the percentage of values in each subgraph. For RoBERTa, the values are normalized across target words, while for ELECTRA, the scores are the raw outputs from its discriminator. }
     \label{fig:dist}
\end{figure*}


\begin{table}[]
\centering
\resizebox{1.0\columnwidth}{!}{%
\begin{tabular}{lcccc} \toprule
               & \tf{COPA} & \tf{StoryCloze} & \tf{HellaSwag} & \tf{PIQA} \\ \midrule
\multicolumn{5}{l}{\textbf{\textit{RoBERTa-base}}} \\
~~~~\CLS  & 69.3 (2.1)   & 81.4 (1.2)      & 32.1 (2.3)      & 54.6 (1.0)   \\
~~~~PET  & \textbf{78.1 (2.6)}    &  69.8 (3.9)     & 30.1 (2.1)      &   61.9 (1.2)         \\
\multicolumn{5}{l}{\textbf{\textit{ELECTRA-base}}} \\
~~~~\CLS  & 76.0 (0.8)   & 84.5 (1.1)      & \textbf{58.8 (2.0)}      & 64.9  (1.1)   \\
~~~~prob & 76.7 (1.7)   & 85.7 (1.5)      & 54.5 (1.0)      & 65.8 (0.3)   \\
~~~~rep  & 76.7 (1.7)   & \textbf{86.6 (1.1)}      & 58.0  (0.7)      & \textbf{66.2 (0.4)}   \\ \midrule
\multicolumn{5}{l}{\textbf{\textit{RoBERTa-large}}} \\
~~~~\CLS  & 76.0  (1.4)   & 86.5  (4.9)      & 44.1  (5.1)      & 55.5 (1.6)   \\
~~~~PET  & 85.4  (2.9)   &        85.4 (5.3)             & 46.9  (5.0)      &      64.6 (3.9)                  \\
\multicolumn{5}{l}{\textbf{\textit{ELECTRA-large}}} \\
~~~~\CLS  & \textbf{96.0 (0.8)}   & \textbf{95.6 (0.7)}      & \textbf{79.9 (2.6)}      & 71.2 (3.8)   \\
~~~~prob & 89.0 (1.6)   & 90.2 (0.5)      & 77.9  (0.6)      & 72.0  (0.7)   \\
~~~~rep  & 88.7 (0.9)   & 90.4 (0.6)      & 78.1  (0.5)      & \textbf{72.7 (1.3)}  \\ \bottomrule
\end{tabular}}
\caption{Multiple-choice task results for prompt-based fine-tuning on RoBERTa and ELECTRA with 32 examples across three runs.  \textit{CLS}, \textit{prob} and \textit{rep} denote that we take the \CLS~representation, the average probability or the average representations for prediction. }
\label{tab:multtoken}
\end{table}

\subsection{Tasks with Single-token Target Words}
\autoref{tab:main} reports zero-shot and few-shot fine-tuning results on base-sized models.\footnote{Results on large-sized models are in \autoref{app:large}.} ELECTRA shows a clear advantage compared to BERT and RoBERTa, with an average margin of 7.9 and 3.5 points on zero-shot prediction, respectively, and an average margin of 10.2 and 3.1 on prompt-based few-shot fine-tuning. The difference is much smaller on standard few-shot fine-tuning (3.1 and 1.1, respectively),\footnote{The gains of ELECTRA over RoBERTa and BERT on full dataset fine-tuning are similar, 3.3 and 1.2, respectively.} suggesting that ELECTRA is inherently better at prompt-based learning, in addition to being a better model in general. On that note, we find that prompt-based fine-tuning consistently outperforms standard fine-tuning in line with prior work \citep{gao2021making, schick2021s}, which reinforces the importance of using prompts in the few-shot learning setting.

\subsection{Tasks with Multi-token Target Options}
For tasks involving multi-token options, we focus on the few-shot fine-tuning setting and we use task-specific templates to encode data in all experiments. For both models, we experiment with the few-shot fine-tuning setting where we map the \CLS~representations to scalars. For RoBERTa, we train a head from scratch and for ELECTRA, we reuse the discriminator head. Additionally, we test the PET approach \cite{schick2021s} on RoBERTa models as illustrated in \autoref{fig:teaser}.

As shown in \autoref{tab:multtoken}, ELECTRA generally presents better and stabler performance than RoBERTa. PET \cite{schick2021s}, which uses a heuristic autoregressive decoding approach, in most cases outperforms RoBERTa with {\CLS} fine-tuning, but still falls behind ELECTRA models. For ELECTRA, using average token representations is comparable or outperforms {\CLS} representations for prediction on the base-sized model but {\CLS} fine-tuning leads to the best performance on the large-sized model.

These results demonstrate the potential of discriminative models on a broader range of tasks under the few-shot setting.\footnote{While we focus on MLMs for their direct comparability, ELECTRA also outperforms GPT-3 results reported in \citet{brown2020language} for equivalent model sizes.}


\section{Analysis}

\begin{figure}[t]
\centering
\includegraphics[width=\linewidth]{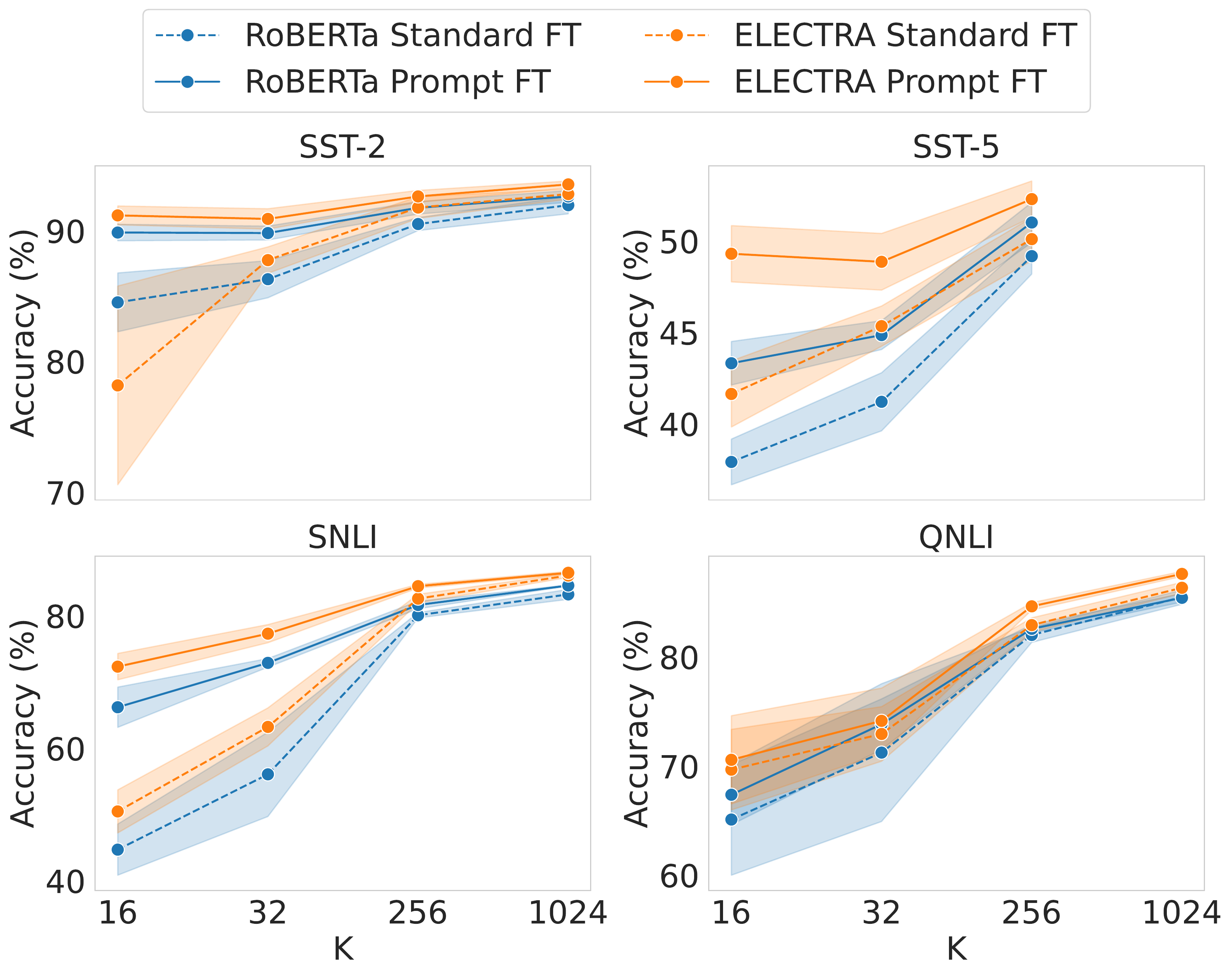}
\caption{Few-shot performance of RoBERTa v.s. ELECTRA with standard and prompt-based fine-tuning as $K$ (\# examples per label) increases. FT: fine-tuning.}
\label{fig:kplot}
\end{figure}

\subsection{Number of Examples}
\autoref{fig:kplot} shows the standard and prompt-based few-shot fine-tuning performance as the number of instances ($K$) increases for RoBERTa and ELECTRA on four datasets.\footnote{See \autoref{app:kplot} for results on the rest of the datasets.} ELECTRA outperforms RoBERTa with a small $K$, and the two converge when $K\ge 256$. The performance gap increases as the number of examples decreases, demonstrating that ELECTRA's discriminative pre-training objective is well-suited for few-shot applications.

\subsection{Prediction Analysis} \autoref{fig:dist} presents the output distributions of zero-shot predictions of RoBERTa and ELECTRA on SST-2.\footnote{In \autoref{app:few}, we show that the output distribution shifts to a polarized shape with few-shot fine-tuning.} We normalize the RoBERTa output across target words (\textit{great}, \textit{terrible}) and keep the ELECTRA output as it is. For negative examples, the predictions from RoBERTa are only slightly skewed towards \textit{terrible}, indicating that RoBERTa likely assigns a similar probability to the antonym \textit{great} when masking the word \textit{terrible}. This finding sheds light on why ELECTRA outperforms RoBERTa, as it has likely seen the closely-related alternative words during training and learned to suppress the probability of these words being original.

We analyze RoBERTa's output distribution on its pre-training corpus to verify that the analysis does not spuriously correlate with the task template. We randomly sample sentences that either contain the word \textit{great} or \textit{terrible} and forward the sentences through the model after masking these two words. We visualize the normalized output distribution over \textit{great} and \textit{terrible} in \autoref{fig:dist} and observe a similar pattern as RoBERTa's zero-shot prediction distribution on SST-2. It corroborates our hypothesis that masked language models fail to predict the correct word but instead output the antonym in some cases, e.g., when the ground truth is \textit{terrible}, which enables ELECTRA to distinguish semantically opposite words and further strengthens its prompt-based prediction ability.

\section{Conclusion}

We explore discriminative pre-trained models for prompt-based zero-shot and few-shot learning. We find that these models consistently outperform masked language models that are trained with equivalent or even less computation, suggesting that discriminative pre-trained models are more effective zero-shot and few-shot learners. Analysis shows that the ELECTRA's generator could very likely feeds negatives like antonyms to the discriminator, which serves as a direct contrast during pre-training. We also speculate that discriminative models are less vulnerable to the surface form competition~\cite{holtzman2021surface}, and we would like to dig deeper into this hypothesis in future work.

\clearpage
\section*{Acknowledgements}

The authors thank Myle Ott, Dan Friedman, Sadhika Malladi, Zexuan Zhong, Tianyu Gao and the anonymous reviewers for their valuable feedback on our paper.

\section*{Limitations}
One limitation of this work is that we limit our exploration within the scope of discriminative tasks. It is prohibitively expensive to apply the prompting approach of ELECTRA to tasks without a limited set of candidates. The prompting approach we propose for ELECTRA requires one forward pass for each option in one example. In contrast, masked language models only require one forward pass for each example. 

Another limitation is that we only include a limited set of continuation-based multiple-choice tasks for evaluation due to space constraints. We leave evaluating on a more diverse set of multiple-option tasks as future work.


\bibliography{anthology, custom}
\bibliographystyle{acl_natbib}

\clearpage
\appendix
\begin{table*}[t]
\resizebox{2\columnwidth}{!}{\begin{tabular}{llcccc} \toprule
Models          & Pretrain Corpora                                       & Corpora Size & \multicolumn{1}{l}{\# Vocab} & \multicolumn{1}{l}{Steps} & \multicolumn{1}{l}{GLUE} \\ \midrule
BERT$_\mathrm{base}$      & Wikipedia, BooksCorpus                                 & 16GB         & 30K                            & 1M                               & 82.2                     \\
RoBERTa$_\mathrm{base}$    & Wikipedia, BooksCorpus, CC-News, OpenWebText, Stores   & 160GB        & 50K                            & 500K                             & 86.4                     \\
ELECTRA$_\mathrm{base}$   & Wikipedia, BooksCorpus                                 & 16GB         & 30K                            & 1M                             & 87.1                     \\ \midrule
BERT$_\mathrm{large}$     & Wikipedia, BooksCorpus                                 & 16GB         & 30K                            & 464K                             & 84.0                       \\
RoBERTA$_\mathrm{large}$ & Wikipedia, BooksCorpus, CC-News, OpenWebText, Stores   & 160GB        & 50K                            & 500K                             & 88.9                     \\
ELECTRA$_\mathrm{large}$  & Wikipedia, BooksCorpus, ClueWeb, CommonCrawl, Gigaword & 33GB         & 30K                            & 400K                             & 89.0    \\ \bottomrule                   
\end{tabular}}
\caption{Pre-training details of BERT, RoBERTa and ELECTRA. The GLUE results are taken from \citet{Clark2020ELECTRA:} and \citet{liu2019roberta} on the development set.}
\label{tab:model}
\end{table*}
\section{Model Details}
\label{app:model}
We list the details of the pre-trained models, including training corpora, vocabulary size, training steps, and GLUE development set results in \autoref{tab:model}. ELECTRA, which is trained on the same set of corpora as BERT, outperforms BERT on GLUE datasets by 3 to 5 points. It slightly underperforms RoBERTa on the base size but is comparable to RoBERTa on the large size.

\section{Datasets}
\label{app:datasets}
We experiment on 1) sentence classification tasks, including 3 sentiment analysis datasets: SST-2, SST-5~\cite{socher2013recursive}, MR~\cite{pang2005seeing}; 4 natural language inference tasks: MNLI~\cite{williams2018broad}, RTE~\cite{dagan2005pascal, haim2006second, giampiccolo2007third, bentivogli2009fifth}, QNLI~\cite{rajpurkar2016squad}, SNLI~\cite{bowman2015large}; AGNews~\cite{zhang2015character}, which is a news classification dataset, BoolQ~\cite{clark-etal-2019-boolq}, which is a dataset of boolean questions; 2) multiple-choice tasks, which involve multi-token options, including COPA~\cite{roemmele2011choice}, StoryCloze~\cite{mostafazadeh2016corpus}, Hellaswag~\cite{zellers2019hellaswag}, PIQA~\cite{bisk2020piqa}. We construct a validation set the same size as the training set in few-shot settings and report results on the full validation set for all datasets.

\section{Training Details}
\label{app:setup}
Following \citet{gao2021making}, we conduct a grid search for all few-shot experiments and take learning rates from \{1e-5, 2e-5, 3e-5\} and batch sizes from $\{2, 4, 8\}$. For each trial, we perform gradients updates for $1000$ steps, evaluate the model every $100$ steps and select the model with the best validation accuracy. For full-shot experiments, we conduct a grid search with learning rates from \{1e-5, 2e-5, 3e-5\} and use a batch size of 16.


\begin{table*}[h!]
\centering
\resizebox{1.6\columnwidth}{!}{
\begin{tabular}{lccc|ccc}\toprule
                     & \multicolumn{3}{c}{\tf{SST-2}}             & \multicolumn{3}{c}{\tf{SST-5}}            \\ \midrule
                     & BERT       & RoBERTa     & ELECTRA    & BERT       & RoBERTa    & ELECTRA    \\ \cmidrule(l{2pt}r{2pt}){2-4} \cmidrule(l{2pt}r{2pt}){5-7}
\cellcolor{ggreen!30}Zero-shot (\cmark)     & 61.2       & 83.6        & \textbf{86.0}       & 25.7       & \textbf{34.7}       & 32.1       \\
\cellcolor{ggreen!60}Few-shot & 82.4 (3.0) & \textbf{85.4 (2.9)}  & 75.8 (5.2) & 40.1 (2.4) & 41.3 (1.2) & \textbf{42.8 (0.9)} \\
\cellcolor{ggreen!60}Few-shot (\cmark)   & 87.9 (0.8) & 93.0 (0.6)  & \textbf{93.6 (0.4)} & 42.4 (1.5) & 47.1 (0.9) & \textbf{50.3 (1.8)} \\
\midrule
\cellcolor{ggreen!90}Full-shot     & {94.3}      & {96.6}       & \textbf{{97.1}}      & {53.3}       & {56.8}      & {\textbf{58.9}}      \\ \midrule
                     & \multicolumn{3}{c}{\tf{SNLI}}              & \multicolumn{3}{c}{\tf{BoolQ}}            \\ \midrule
                     & BERT       & RoBERTa     & ELECTRA    & BERT       & RoBERTa    & ELECTRA    \\ \cmidrule(l{2pt}r{2pt}){2-4} \cmidrule(l{2pt}r{2pt}){5-7}
\cellcolor{ggreen!30}Zero-shot (\cmark)     & 41.5       & 49.8        & \textbf{59.4}       & 49.3       & 53.4       & \textbf{71.1}       \\
\cellcolor{ggreen!60}Few-shot  & 51.2 (3.3) & 51.4 (3.1)  & \textbf{66.7 (2.7)} & 56.0 (2.3) & 59.5 (3.0) & \textbf{61.3 (1.5)} \\
\cellcolor{ggreen!60}Few-shot (\cmark)  & 60.6 (2.8) & \textbf{79.4 (1.4)} & 79.1 (2.0) & 56.9 (0.3) & 70.3 (2.6) & \textbf{75.2 (1.2)} \\
\midrule
\cellcolor{ggreen!90}Full-shot     & {91.6}      & {92.1}       & \textbf{{92.2}}      &      {73.1}      & \textbf{{85.2}}       &    {85.0}       \\ \bottomrule
\end{tabular}
}
\caption{Zero-shot and few-shot (16 examples per label) and full-shot results of large-sized BERT, RoBERTa and ELECTRA. \cmark: denotes whether prompts are used or not. }
\label{tab:large}
\end{table*}

\section{Results on Large-sized Models}
\label{app:large}
We present prompt-based zero-shot and few-shot results on large-sized models in \autoref{tab:large} to show that the trend prevails when the model scales up.  Except for SNLI, the average gain from prompt-based fine-tuning for ELECTRA is significantly larger than BERT and RoBERTa. Notably,  ELECTRA also significantly outperforms BERT and RoBERTa on zero-shot prediction.

\section{Number of Examples}
\label{app:kplot}
\begin{figure}[!htbp]
\centering
\includegraphics[width=\linewidth]{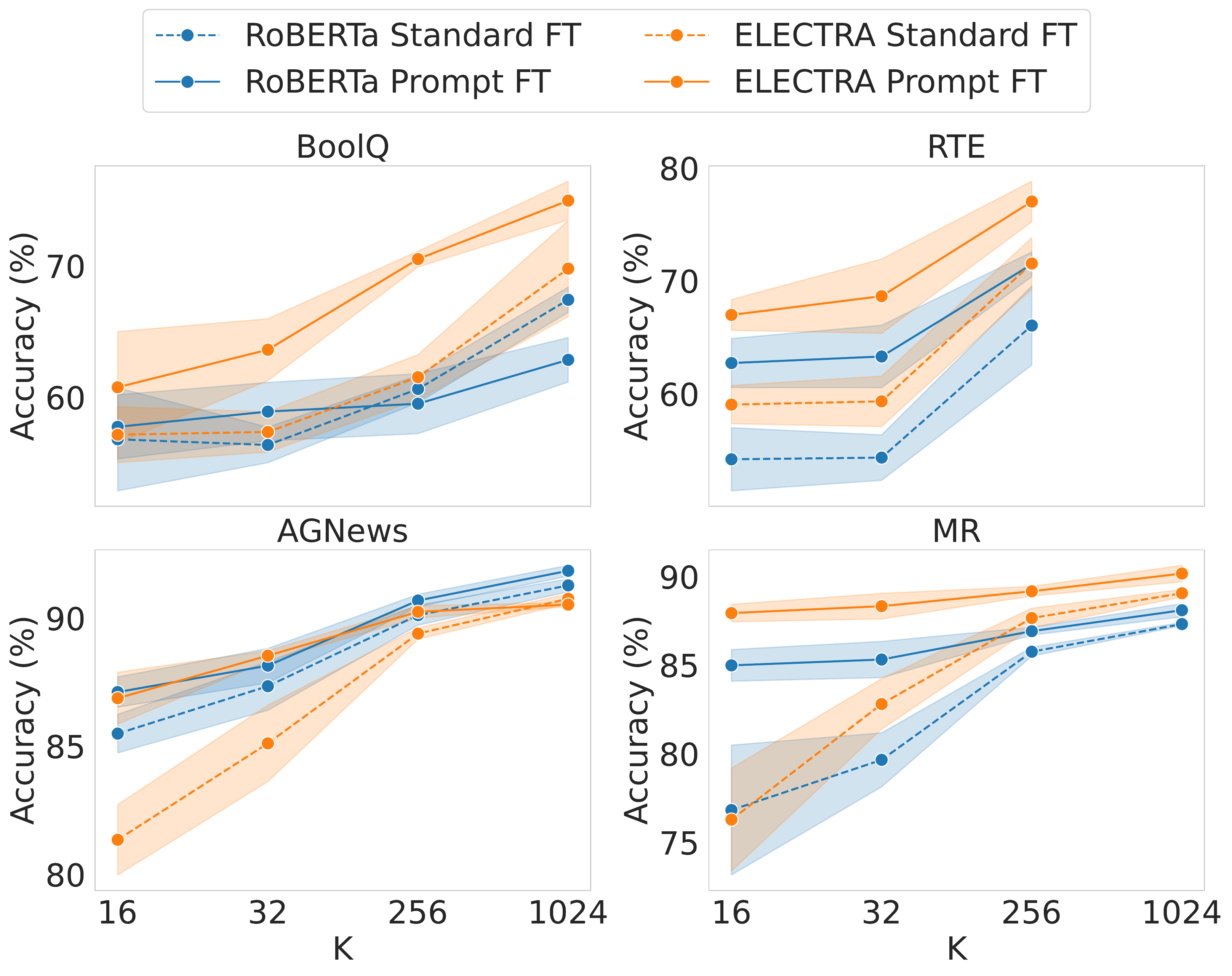}
\caption{Few-shot performance of RoBERTa v.s. ELECTRA with standard and prompt-based fine-tuning as K (the number of instances per label) increases on more tasks.}
\label{fig:kplot2}
\end{figure}

We show the few-shot results as a function of $K$ on BoolQ, RTE, AGNews and MR in \autoref{fig:kplot2}.
ELECTRA significantly outperforms RoBERTa on BoolQ and RTE across all settings, suggesting that ELECTRA is an overall stronger model for these datasets. On MR, we observe a similar pattern where the gap between ELECTRA and RoBERTa gets smaller, showing that ELECTRA benefits from prompt training more than RoBERTa. On AGNews, ELECTRA underperforms RoBERTa on standard fine-tuning but closes the gap on prompt-based fine-tuning, backing up the argument that ELECTRA benefits more from the prompt.


\begin{table*}[t]
\centering
\resizebox{1.25\columnwidth}{!}{\begin{tabular}{lrccc} \toprule
\textbf{Task}  & $\mathbf{K}$  & \textbf{Original}                   & \textbf{Original w/o shuffling} & \textbf{Contrastive} \\ \midrule
\multirow{4}{*}{SST-2}  & 16                    & \tf{91.2} (0.7)                     & \tf{91.2} (0.8)    & 91.0 (0.4)  \\
                        & 32                    & \tf{90.9} (0.8)                     & 90.5 (0.8)    & 90.6 (0.7)  \\
                        & 256                   & \tf{92.6} (0.5)                     & 92.2 (0.4)    & 92.2 (0.7)  \\
                        & 1024                  & \tf{93.6} (0.3)                     & 92.9 (0.5)    & 93.1 (0.3)  \\ \midrule
\multirow{4}{*}{AGNews} & 16                    & \multicolumn{1}{c}{\tf{86.5} (1.1)} & 85.4 (1.3)    & 85.4 (0.8)  \\
                        & 32                    & \multicolumn{1}{c}{\tf{88.4} (0.3)} & 86.5 (0.6)    & 86.7 (0.7)  \\
                        & 256                   & \multicolumn{1}{c}{\tf{90.3} (0.2)} & 89.8 (0.2)    & 89.3 (0.2)  \\
                        & 1024                  & \multicolumn{1}{c}{\tf{90.5} (0.1)} & 90.1 (0.2)    & 89.5 (0.3) \\ \bottomrule
\end{tabular}}
\caption{Few-shot prompt-based fine-tuning results on different objectives with ELECTRA${_\text{base}}$. Original w/o shuffling denotes that we load the batches without data shuffling to mimic the data loading restriction when training with the contrastive objective).} 
\label{tab:contrastive}
\end{table*}

\section{An Alternative Contrastive Objective}
\label{app:con}
We also explored another contrastive objective with ELECTRA's logits for prompt-based few-shot finetuning. For all the prompts of an input $x$ with the label set $\mathcal{Y}$, we define the loss as
\begin{equation}
- \log \frac{\exp (\phi(c(y)))}{\sum_{y' \in \mathcal{Y}} \exp (\phi(c(y'))))} \nonumber
\end{equation}
\noindent where $\mathcal{H}(x) = \frac{1}{1+e^{-\phi(x)}}$ and $\phi(x)$ denote the logits from the discriminator. We directly contrast the correct target option with the incorrect ones with this objective. We show results on SST-2 and AGNews in \autoref{tab:contrastive}. Prompt-based fine-tuning with the original ELECTRA objective outperforms the contrastive objective. We hypothesize that the downside of the contrastive objective is that it forces one input with different target options to be packed into the same batch instead of shuffling the whole dataset randomly, and it affects the optimization. We also experiment on the original discriminative objective with the same batch restriction and observe a performance drop to verify the hypothesis.

\section{Few-shot Output Distribution}
\label{app:few}
\begin{figure*}[h]
     \centering
     \begin{subfigure}[b]{0.4\textwidth}
         \centering
         \includegraphics[width=\textwidth]{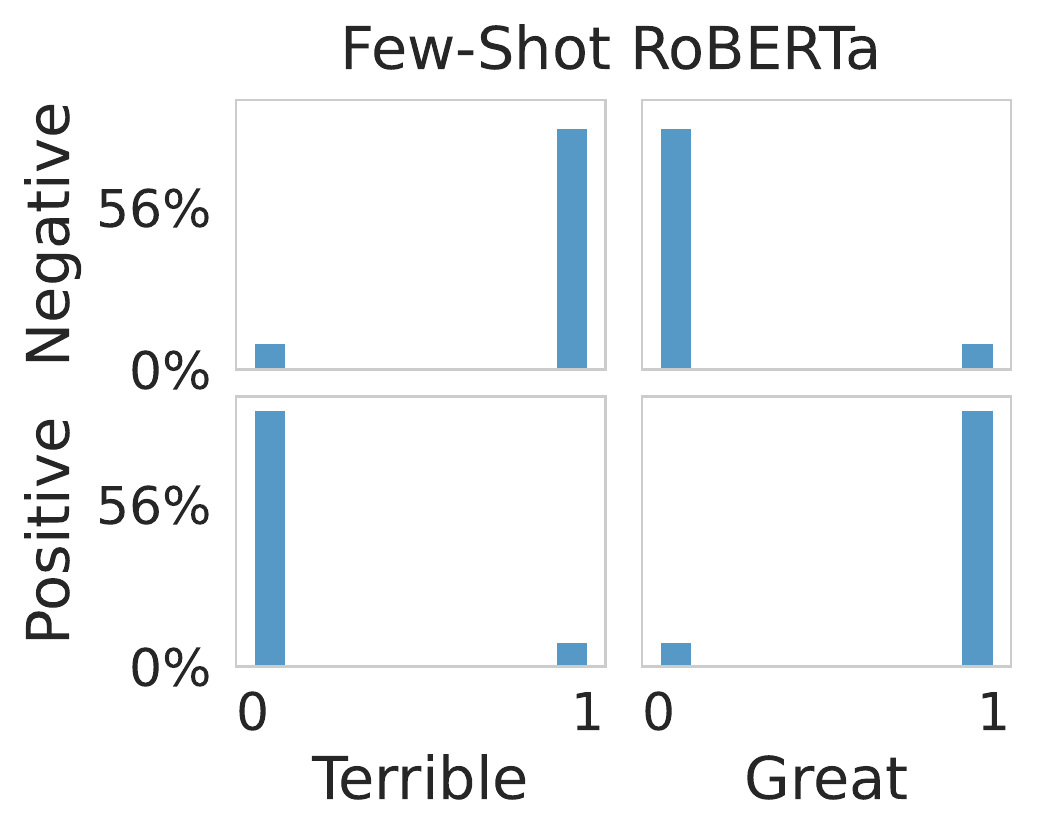}
     \end{subfigure}
     \begin{subfigure}[b]{0.4\textwidth}
         \centering
         \includegraphics[width=\textwidth]{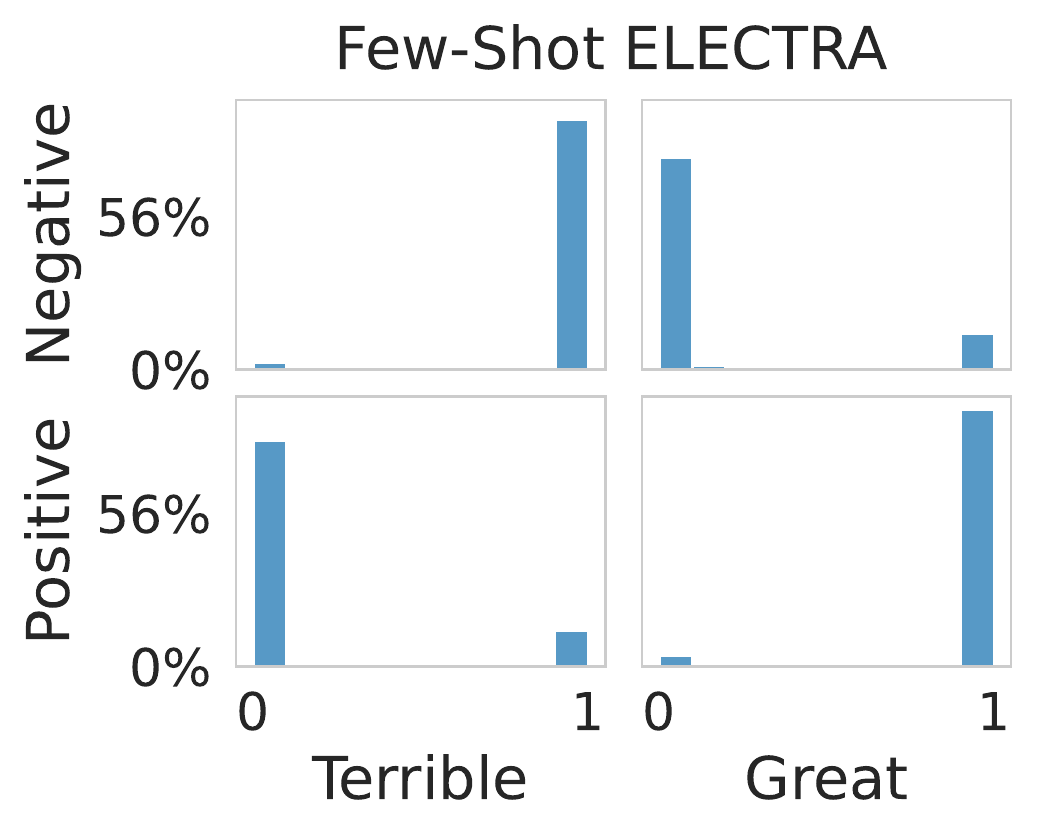}
     \end{subfigure}
     \caption{Few-shot prediction distributions on SST-2 with RoBERTa$_\text{base}$ and ELECTRA$_\text{base}$. Each sub-graph shows the output distribution for inputs with a label $y \in \{\text{negative}, \text{positive}\}$ when prompted with the corresponding target option $\mathcal{M}(y)$. }
     \label{fig:dist2}
\end{figure*}

We show the few-shot output distribution of RoBERTa and ELECTRA on SST-2 in \autoref{fig:dist2}. The output distributions are polarized after few-shot training.

\section{Prompts}
\label{app:template}
\begin{table*}[t]
\centering
\resizebox{1.8\columnwidth}{!}{%
\begin{tabular}{lll} \toprule 
\textbf{Task} & \textbf{Template} & \textbf{Label Words} \\ \midrule
SST-2      & \textless{}sentence\textgreater{}  It was {\texttt{[MASK]}} .                                                                                               & positive: great, negative: terrible                                                   \\
SST-5      & \textless{}sentence\textgreater{}  It was {\texttt{[MASK]}}  .                                                                                               & v.positive: great, positive: good, neutral: okay, \\
& & negative: bad, v.negative: terrible \\
MR         & \textless{}sentence\textgreater{}  It was {\texttt{[MASK]}}  .                                                                                               & positive: great, negative: terrible                                                   \\
MNLI       & \textless{}premise\textgreater{}? {\texttt{[MASK]}}  , \textless{}hypothesis\textgreater{}                                                                & entailment: Yes, neutral: Maybe, contradiction: No                                    \\
SNLI       & \textless{}premise\textgreater{}? {\texttt{[MASK]}}  , \textless{}hypothesis\textgreater{}                                                                & entailment: Yes, neutral: Maybe, contradiction: No                                    \\
RTE        & \textless{}premise\textgreater{}? {\texttt{[MASK]}}  , \textless{}hypothesis\textgreater{}                                                                & entailment: Yes, not entailment: No                                                   \\
QNLI       & \textless{}premise\textgreater{}? {\texttt{[MASK]}}  , \textless{}hypothesis\textgreater{}                                                                & entailment: Yes, not entailment: No                                                   \\
AGNews     & {\texttt{[MASK]}}  News: \textless{}sentence\textgreater{}                                                                                                & World: World, Sports: Sports, \\ 
& & Business: Business, Sci/Tech: Tech                    \\ 
BoolQ      & \textless passage\textgreater{} Question: \textless{}question\textgreater{} ? Answer: {\texttt{[MASK]}}  .                                                  & No: No, Yes: Yes                                                                      \\ 
\bottomrule
\end{tabular}}
\caption{Task templates for tasks with single-token verbalizers.}
\label{tab:single-template}
\end{table*}

\begin{table*}[t]
\centering
\resizebox{1.4\columnwidth}{!}{%
\begin{tabular}{ll} \toprule
\textbf{Task} & \textbf{Template}  \\ \midrule
COPA       & \textless{}sentence\textgreater{} so/because{}  {\texttt{[OPTION]}}                                                                                     \\
StoryCloze & \textless{}sentence1\textgreater{} \textless{}sentence2\textgreater{} \textless{}sentence3\textgreater{} \textless{}sentence4\textgreater{} {\texttt{[OPTION]}}                                                                                        \\
Hellaswag  & \textless{}context\textgreater {\texttt{[OPTION]}}                                                                                                                                                                                             \\
PIQA       & \textless{}sentence\textgreater {\texttt{[OPTION]}}   \\ \bottomrule                                                                                                                                                                         
\end{tabular}}
\caption{Task templates for tasks with multi-token verbalizers.}
\label{tab:multi-template}
\end{table*}

We largely follow previous works to construct our prompts. For sentiment classification tasks and natural language inference tasks, we use prompts from~\citet{gao2021making}. For AGNews, we use the prompt from~\citet{holtzman2021surface} and for BoolQ, we use the prompt from~\citet{schick2021s}. For tasks involving multi-token options, we simply concatenate the context and options, which largely follows~\citet{holtzman2021surface}. The prompt details can be found in \autoref{tab:single-template} and \autoref{tab:multi-template}.

To verify that the prompts does not affect our major conclusion, we conduct prompt-based few-shot finetuning experiments with different prompts for four tasks. The prompts we use are in \autoref{tab:tem_sen}. Results in \autoref{tab:template_sensitivity} show that ELECTRA outperforms RoBERTa with different prompts.

\begin{table*}[h]
\resizebox{2.0\columnwidth}{!}{%
\begin{tabular}{lll} \toprule 
\textbf{Text} & $\mathcal{T}$& \textbf{Template} \\ \midrule
\multirow{3}{*}{MNLI}       & $\mathcal{T}_1$ & \textless{}premise\textgreater{} ? {\texttt{[MASK]}} , \textless{}hypothesis\textgreater{}                                                                                                    \\  
                            & $\mathcal{T}_2$ & \textless{}premise\textgreater{} ? {\texttt{[MASK]}} . \textless{}hypothesis\textgreater{}                                                                                                    \\
                            & $\mathcal{T}_3$ & "\textless{}premise\textgreater{}"{} ? {\texttt{[MASK]}} , "\textless{}hypothesis\textgreater{}"                                                                                                \\ \midrule
\multirow{3}{*}{RTE}        & $\mathcal{T}_1$ & \textless{}premise\textgreater{} ? {\texttt{[MASK]}} , \textless{}hypothesis\textgreater{}                                                                                                    \\
                            & $\mathcal{T}_2$ & \textless{}premise\textgreater{} ? {\texttt{[MASK]}} . \textless{}hypothesis\textgreater{}                                                                                                    \\
                            & $\mathcal{T}_3$ & "\textless{}premise\textgreater{}"{} ? {\texttt{[MASK]}} , "\textless{}hypothesis\textgreater{}"                                                                                                \\ \midrule
\multirow{2}{*}{COPA}       & $\mathcal{T}_1$ & \textless{}sentence\textgreater{}  so/because {\texttt{[OPTION]}}                                                                                                                               \\
                            & $\mathcal{T}_2$ & {\texttt{[OPTION\_1]}} or {\texttt{[OPTION\_2]}} ? \textless{}sentence\textgreater so/because {\texttt{[OPTION]}}                                                                                           \\ \midrule
\multirow{2}{*}{StoryCloze} & $\mathcal{T}_1$ & \textless{}sentence1\textgreater{}  \textless{} sentence2\textgreater{}  \textless{} sentence3\textgreater{}  \textless{} sentence4\textgreater{}   {\texttt{[OPTION]}}                                     \\
                            & $\mathcal{T}_2$ & {\texttt{[OPTION\_1]}} or {\texttt{[OPTION\_2]}} ? \textless{}sentence1\textgreater{}  \textless{}sentence2\textgreater{}  \textless{}sentence3\textgreater{}  \textless{}sentence4\textgreater{}  {\texttt{[OPTION]}} \\ \bottomrule
\end{tabular}}
\caption{Task templates for task sensitivity test.}

\label{tab:tem_sen}
\end{table*}

\begin{table*}[t]
\vspace{-30em}
\centering
\resizebox{1.0\columnwidth}{!}{%
\begin{tabular}{clcccc} \toprule
&         & MNLI & RTE  & COPA & SC \\ \midrule
\multirow{2}{*}{$\mathcal{T}_1$} & RoBERTa & 59.1 & 62.7 & 72.7 & 71.0       \\
                    & ELECTRA & \tf{60.8} & \tf{67.0} & \tf{75.0} & \tf{86.9}       \\ \midrule
\multirow{2}{*}{$\mathcal{T}_2$} & RoBERTa & 55.3 & 63.2 & 69.7 & 71.7       \\
                    & ELECTRA & \tf{61.0} & \tf{64.9} & \tf{74.7} & \tf{86.4}       \\ \midrule
\multirow{2}{*}{$\mathcal{T}_3$} & RoBERTa & 57.3 & 63.9 & -    & -          \\
                    & ELECTRA & \tf{60.9} & \tf{67.2} & -    & -   \\ \bottomrule   
\end{tabular}}
\caption{Few-shot results with different templates with base-sized models. ELECTRA still outperforms RoBERTa with different templates (provided in \autoref{tab:tem_sen}). }
\label{tab:template_sensitivity}
\end{table*}

\end{document}